\documentclass[conference,10pt,twocolumn,letter]{IEEEtran}

\usepackage[utf8]{inputenc} % allow utf-8 input
\usepackage[T1]{fontenc}    % use 8-bit T1 fonts
\usepackage{url}            % simple URL typesetting
\usepackage{graphicx}
\usepackage{nicefrac}       % compact symbols for 1/2, etc.
\usepackage{amssymb}
\usepackage{amsmath}
\usepackage{amsthm}
\usepackage{paralist}
\usepackage{setspace}
\usepackage{booktabs}
\usepackage{algorithm}
\usepackage{microtype}
\usepackage{cite}
\usepackage{algpseudocode}
\usepackage{array}
\usepackage{balance}
\usepackage{multirow}
\usepackage{url}            % simple URL typesetting

\usepackage[labelformat=simple,font={footnotesize,sf}]{subcaption}

\usepackage{amssymb}
\usepackage{amsfonts}
\usepackage{mathrsfs}
\usepackage{xspace}
\usepackage{bm}
\usepackage{upgreek}

\newcommand{\safemath}[2]{\newcommand{#1}{\ensuremath{#2}\xspace}}

\safemath{\bma}{\mathbf{a}}
\safemath{\bmb}{\mathbf{b}}
\safemath{\bmc}{\mathbf{c}}
\safemath{\bmd}{\mathbf{d}}
\safemath{\bme}{\mathbf{e}}
\safemath{\bmf}{\mathbf{f}}
\safemath{\bmg}{\mathbf{g}}
\safemath{\bmh}{\mathbf{h}}
\safemath{\bmi}{\mathbf{i}}
\safemath{\bmj}{\mathbf{j}}
\safemath{\bmk}{\mathbf{k}}
\safemath{\bml}{\mathbf{l}}
\safemath{\bmm}{\mathbf{m}}
\safemath{\bmn}{\mathbf{n}}
\safemath{\bmo}{\mathbf{o}}
\safemath{\bmp}{\mathbf{p}}
\safemath{\bmq}{\mathbf{q}}
\safemath{\bmr}{\mathbf{r}}
\safemath{\bms}{\mathbf{s}}
\safemath{\bmt}{\mathbf{t}}
\safemath{\bmu}{\mathbf{u}}
\safemath{\bmv}{\mathbf{v}}
\safemath{\bmw}{\mathbf{w}}
\safemath{\bmx}{\mathbf{x}}
\safemath{\bmy}{\mathbf{y}}
\safemath{\bmz}{\mathbf{z}}
\safemath{\bmzero}{\mathbf{0}}
\safemath{\bmone}{\mathbf{1}}

\bmdefine{\biad}{a}
\bmdefine{\bibd}{b}
\bmdefine{\bicd}{c}
\bmdefine{\bidd}{d}
\bmdefine{\bied}{e}
\bmdefine{\bifd}{f}
\bmdefine{\bigd}{g}
\bmdefine{\bihd}{h}
\bmdefine{\biid}{i}
\bmdefine{\bijd}{j}
\bmdefine{\bikd}{k}
\bmdefine{\bild}{l}
\bmdefine{\bimd}{m}
\bmdefine{\bind}{n}
\bmdefine{\biod}{o}
\bmdefine{\bipd}{p}
\bmdefine{\biqd}{q}
\bmdefine{\bird}{r}
\bmdefine{\bisd}{s}
\bmdefine{\bitd}{t}
\bmdefine{\biud}{u}
\bmdefine{\bivd}{v}
\bmdefine{\biwd}{w}
\bmdefine{\bixd}{x}
\bmdefine{\biyd}{y}
\bmdefine{\bizd}{z}

\bmdefine{\bixid}{\xi}
\bmdefine{\bilambdad}{\lambda}
\bmdefine{\bimud}{\mu}
\bmdefine{\bithetad}{\theta}
\bmdefine{\biphid}{\phi}
\bmdefine{\bideltad}{\delta}

\safemath{\bmia}{\biad}
\safemath{\bmib}{\bibd}
\safemath{\bmic}{\bicd}
\safemath{\bmid}{\bidd}
\safemath{\bmie}{\bied}
\safemath{\bmif}{\bifd}
\safemath{\bmig}{\bigd}
\safemath{\bmih}{\bihd}
\safemath{\bmii}{\biid}
\safemath{\bmij}{\bijd}
\safemath{\bmik}{\bikd}
\safemath{\bmil}{\bild}
\safemath{\bmim}{\bimd}
\safemath{\bmin}{\bind}
\safemath{\bmio}{\biod}
\safemath{\bmip}{\bipd}
\safemath{\bmiq}{\biqd}
\safemath{\bmir}{\bird}
\safemath{\bmis}{\bisd}
\safemath{\bmit}{\bitd}
\safemath{\bmiu}{\biud}
\safemath{\bmiv}{\bivd}
\safemath{\bmiw}{\biwd}
\safemath{\bmix}{\bixd}
\safemath{\bmiy}{\biyd}
\safemath{\bmiz}{\bizd}

\safemath{\bmxi}{\bixid}
\safemath{\bmlambda}{\bilambdad}
\safemath{\bmmu}{\bimud}
\safemath{\bmtheta}{\bithetad}
\safemath{\bmphi}{\biphid}
\safemath{\bmdelta}{\bideltad}

\safemath{\bA}{\mathbf{A}}
\safemath{\bB}{\mathbf{B}}
\safemath{\bC}{\mathbf{C}}
\safemath{\bD}{\mathbf{D}}
\safemath{\bE}{\mathbf{E}}
\safemath{\bF}{\mathbf{F}}
\safemath{\bG}{\mathbf{G}}
\safemath{\bH}{\mathbf{H}}
\safemath{\bI}{\mathbf{I}}
\safemath{\bJ}{\mathbf{J}}
\safemath{\bK}{\mathbf{K}}
\safemath{\bL}{\mathbf{L}}
\safemath{\bM}{\mathbf{M}}
\safemath{\bN}{\mathbf{N}}
\safemath{\bO}{\mathbf{O}}
\safemath{\bP}{\mathbf{P}}
\safemath{\bQ}{\mathbf{Q}}
\safemath{\bR}{\mathbf{R}}
\safemath{\bS}{\mathbf{S}}
\safemath{\bT}{\mathbf{T}}
\safemath{\bU}{\mathbf{U}}
\safemath{\bV}{\mathbf{V}}
\safemath{\bW}{\mathbf{W}}
\safemath{\bX}{\mathbf{X}}
\safemath{\bY}{\mathbf{Y}}
\safemath{\bZ}{\mathbf{Z}}

\safemath{\bZero}{\mathbf{0}}
\safemath{\bOne}{\mathbf{1}}
\safemath{\bDelta}{\mathbf{\Delta}}
\safemath{\bLambda}{\boldsymbol\Lambda}
\safemath{\bPhi}{\mathbf{\Upphi}}
\safemath{\bSigma}{\mathbf{\Upsigma}}
\safemath{\bOmega}{\mathbf{\Upomega}}
\safemath{\bTheta}{\mathbf{\Uptheta}}

\bmdefine{\biAd}{A}
\bmdefine{\biBd}{B}
\bmdefine{\biCd}{C}
\bmdefine{\biDd}{D}
\bmdefine{\biEd}{E}
\bmdefine{\biFd}{F}
\bmdefine{\biGd}{G}
\bmdefine{\biHd}{H}
\bmdefine{\biId}{I}
\bmdefine{\biJd}{J}
\bmdefine{\biKd}{K}
\bmdefine{\biLd}{L}
\bmdefine{\biMd}{M}
\bmdefine{\biOd}{N}
\bmdefine{\biPd}{O}
\bmdefine{\biQd}{P}
\bmdefine{\biRd}{R}
\bmdefine{\biSd}{S}
\bmdefine{\biTd}{T}
\bmdefine{\biUd}{U}
\bmdefine{\biVd}{V}
\bmdefine{\biWd}{W}
\bmdefine{\biXd}{X}
\bmdefine{\biYd}{Y}
\bmdefine{\biZd}{Z}

\bmdefine{\biDelta}{\Delta}
\bmdefine{\biLambda}{\Lambda}
\bmdefine{\biPhi}{\Phi}
\bmdefine{\biSigma}{\Sigma}
\bmdefine{\biOmega}{\Omega}
\bmdefine{\biTheta}{\Theta}

\safemath{\bimA}{\biAd}
\safemath{\bimB}{\biBd}
\safemath{\bimC}{\biCd}
\safemath{\bimD}{\biDd}
\safemath{\bimE}{\biEd}
\safemath{\bimF}{\biFd}
\safemath{\bimG}{\biGd}
\safemath{\bimH}{\biHd}
\safemath{\bimI}{\biId}
\safemath{\bimJ}{\biJd}
\safemath{\bimK}{\biKd}
\safemath{\bimL}{\biLd}
\safemath{\bimM}{\biMd}
\safemath{\bimN}{\biNd}
\safemath{\bimO}{\biOd}
\safemath{\bimP}{\biPd}
\safemath{\bimQ}{\biQd}
\safemath{\bimR}{\biRd}
\safemath{\bimS}{\biSd}
\safemath{\bimT}{\biTd}
\safemath{\bimU}{\biUd}
\safemath{\bimV}{\biVd}
\safemath{\bimW}{\biWd}
\safemath{\bimX}{\biXd}
\safemath{\bimY}{\biYd}
\safemath{\bimZ}{\biZd}

\safemath{\bimDelta}{\biDelta}
\safemath{\bimLambda}{\biLambda}
\safemath{\bimPhi}{\biPhi}
\safemath{\bimSigma}{\biSigma}
\safemath{\bimOmega}{\biOmega}
\safemath{\bimTheta}{\biTheta}

\safemath{\setA}{\mathcal{A}}
\safemath{\setB}{\mathcal{B}}
\safemath{\setC}{\mathcal{C}}
\safemath{\setD}{\mathcal{D}}
\safemath{\setE}{\mathcal{E}}
\safemath{\setF}{\mathcal{F}}
\safemath{\setG}{\mathcal{G}}
\safemath{\setH}{\mathcal{H}}
\safemath{\setI}{\mathcal{I}}
\safemath{\setJ}{\mathcal{J}}
\safemath{\setK}{\mathcal{K}}
\safemath{\setL}{\mathcal{L}}
\safemath{\setM}{\mathcal{M}}
\safemath{\setN}{\mathcal{N}}
\safemath{\setO}{\mathcal{O}}
\safemath{\setP}{\mathcal{P}}
\safemath{\setQ}{\mathcal{Q}}
\safemath{\setR}{\mathcal{R}}
\safemath{\setS}{\mathcal{S}}
\safemath{\setT}{\mathcal{T}}
\safemath{\setU}{\mathcal{U}}
\safemath{\setV}{\mathcal{V}}
\safemath{\setW}{\mathcal{W}}
\safemath{\setX}{\mathcal{X}}
\safemath{\setY}{\mathcal{Y}}
\safemath{\setZ}{\mathcal{Z}}
\safemath{\emptySet}{\varnothing}

\safemath{\colA}{\mathscr{A}}
\safemath{\colB}{\mathscr{B}}
\safemath{\colC}{\mathscr{C}}
\safemath{\colD}{\mathscr{D}}
\safemath{\colE}{\mathscr{E}}
\safemath{\colF}{\mathscr{F}}
\safemath{\colG}{\mathscr{G}}
\safemath{\colH}{\mathscr{H}}
\safemath{\colI}{\mathscr{I}}
\safemath{\colJ}{\mathscr{J}}
\safemath{\colK}{\mathscr{K}}
\safemath{\colL}{\mathscr{L}}
\safemath{\colM}{\mathscr{M}}
\safemath{\colN}{\mathscr{N}}
\safemath{\colO}{\mathscr{O}}
\safemath{\colP}{\mathscr{P}}
\safemath{\colQ}{\mathscr{Q}}
\safemath{\colR}{\mathscr{R}}
\safemath{\colS}{\mathscr{S}}
\safemath{\colT}{\mathscr{T}}
\safemath{\colU}{\mathscr{U}}
\safemath{\colV}{\mathscr{V}}
\safemath{\colW}{\mathscr{W}}
\safemath{\colX}{\mathscr{X}}
\safemath{\colY}{\mathscr{Y}}
\safemath{\colZ}{\mathscr{Z}}

\safemath{\opA}{\mathbb{A}}
\safemath{\opB}{\mathbb{B}}
\safemath{\opC}{\mathbb{C}}
\safemath{\opD}{\mathbb{D}}
\safemath{\opE}{\mathbb{E}}
\safemath{\opF}{\mathbb{F}}
\safemath{\opG}{\mathbb{G}}
\safemath{\opH}{\mathbb{H}}
\safemath{\opI}{\mathbb{I}}
\safemath{\opJ}{\mathbb{J}}
\safemath{\opK}{\mathbb{K}}
\safemath{\opL}{\mathbb{L}}
\safemath{\opM}{\mathbb{M}}
\safemath{\opN}{\mathbb{N}}
\safemath{\opO}{\mathbb{O}}
\safemath{\opP}{\mathbb{P}}
\safemath{\opQ}{\mathbb{Q}}
\safemath{\opR}{\mathbb{R}}
\safemath{\opS}{\mathbb{S}}
\safemath{\opT}{\mathbb{T}}
\safemath{\opU}{\mathbb{U}}
\safemath{\opV}{\mathbb{V}}
\safemath{\opW}{\mathbb{W}}
\safemath{\opX}{\mathbb{X}}
\safemath{\opY}{\mathbb{Y}}
\safemath{\opZ}{\mathbb{Z}}
\safemath{\opZero}{\mathbb{O}}
\safemath{\identityop}{\opI}

\safemath{\veca}{\bma}
\safemath{\vecb}{\bmb}
\safemath{\vecc}{\bmc}
\safemath{\vecd}{\bmd}
\safemath{\vece}{\bme}
\safemath{\vecf}{\bmf}
\safemath{\vecg}{\bmg}
\safemath{\vech}{\bmh}
\safemath{\veci}{\bmi}
\safemath{\vecj}{\bmj}
\safemath{\veck}{\bmk}
\safemath{\vecl}{\bml}
\safemath{\vecm}{\bmm}
\safemath{\vecn}{\bmn}
\safemath{\veco}{\bmo}
\safemath{\vecp}{\bmp}
\safemath{\vecq}{\bmq}
\safemath{\vecr}{\bmr}
\safemath{\vecs}{\bms}
\safemath{\vect}{\bmt}
\safemath{\vecu}{\bmu}
\safemath{\vecv}{\bmv}
\safemath{\vecw}{\bmw}
\safemath{\vecx}{\bmx}
\safemath{\vecy}{\bmy}
\safemath{\vecz}{\bmz}

\safemath{\veczero}{\bmzero}
\safemath{\vecone}{\bmone}
\safemath{\vecxi}{\bmxi}
\safemath{\veclambda}{\bmlambda}
\safemath{\vecmu}{\bmmu}
\safemath{\vectheta}{\bmtheta}
\safemath{\vecphi}{\bmphi}
\safemath{\vecdelta}{\bmdelta}

\safemath{\matA}{\bA}
\safemath{\matB}{\bB}
\safemath{\matC}{\bC}
\safemath{\matD}{\bD}
\safemath{\matE}{\bE}
\safemath{\matF}{\bF}
\safemath{\matG}{\bG}
\safemath{\matH}{\bH}
\safemath{\matI}{\bI}
\safemath{\matJ}{\bJ}
\safemath{\matK}{\bK}
\safemath{\matL}{\bL}
\safemath{\matM}{\bM}
\safemath{\matN}{\bN}
\safemath{\matO}{\bO}
\safemath{\matP}{\bP}
\safemath{\matQ}{\bQ}
\safemath{\matR}{\bR}
\safemath{\matS}{\bS}
\safemath{\matT}{\bT}
\safemath{\matU}{\bU}
\safemath{\matV}{\bV}
\safemath{\matW}{\bW}
\safemath{\matX}{\bX}
\safemath{\matY}{\bY}
\safemath{\matZ}{\bZ}
\safemath{\matzero}{\bmzero}

\safemath{\matDelta}{\bDelta}
\safemath{\matLambda}{\bLambda}
\safemath{\matPhi}{\bPhi}
\safemath{\matSigma}{\bSigma}
\safemath{\matOmega}{\bOmega}
\safemath{\matTheta}{\bTheta}

\safemath{\matidentity}{\matI}
\safemath{\matone}{\matO}

\safemath{\rnda}{A}
\safemath{\rndb}{B}
\safemath{\rndc}{C}
\safemath{\rndd}{D}
\safemath{\rnde}{E}
\safemath{\rndf}{F}
\safemath{\rndg}{G}
\safemath{\rndh}{H}
\safemath{\rndi}{I}
\safemath{\rndj}{J}
\safemath{\rndk}{K}
\safemath{\rndl}{L}
\safemath{\rndm}{M}
\safemath{\rndn}{N}
\safemath{\rndo}{O}
\safemath{\rndp}{P}
\safemath{\rndq}{Q}
\safemath{\rndr}{R}
\safemath{\rnds}{S}
\safemath{\rndt}{T}
\safemath{\rndu}{U}
\safemath{\rndv}{V}
\safemath{\rndw}{W}
\safemath{\rndx}{X}
\safemath{\rndy}{Y}
\safemath{\rndz}{Z}

\safemath{\rveca}{\bimA}
\safemath{\rvecb}{\bimB}
\safemath{\rvecc}{\bimC}
\safemath{\rvecd}{\bimD}
\safemath{\rvece}{\bimE}
\safemath{\rvecf}{\bimF}
\safemath{\rvecg}{\bimG}
\safemath{\rvech}{\bimH}
\safemath{\rveci}{\bimI}
\safemath{\rvecj}{\bimJ}
\safemath{\rveck}{\bimK}
\safemath{\rvecl}{\bimL}
\safemath{\rvecm}{\bimM}
\safemath{\rvecn}{\bimN}
\safemath{\rveco}{\bomO}
\safemath{\rvecp}{\bimP}
\safemath{\rvecq}{\bimQ}
\safemath{\rvecr}{\bimR}
\safemath{\rvecs}{\bimS}
\safemath{\rvect}{\bimT}
\safemath{\rvecu}{\bimU}
\safemath{\rvecv}{\bimV}
\safemath{\rvecw}{\bimW}
\safemath{\rvecx}{\bimX}
\safemath{\rvecy}{\bimY}
\safemath{\rvecz}{\bimZ}

\safemath{\rvecxi}{\bmxi}
\safemath{\rveclambda}{\bmlambda}
\safemath{\rvecmu}{\bmmu}
\safemath{\rvectheta}{\bmtheta}
\safemath{\rvecphi}{\bmphi}

\safemath{\rmatA}{\bimA}
\safemath{\rmatB}{\bimB}
\safemath{\rmatC}{\bimC}
\safemath{\rmatD}{\bimD}
\safemath{\rmatE}{\bimE}
\safemath{\rmatF}{\bimF}
\safemath{\rmatG}{\bimG}
\safemath{\rmatH}{\bimH}
\safemath{\rmatI}{\bimI}
\safemath{\rmatJ}{\bimJ}
\safemath{\rmatK}{\bimK}
\safemath{\rmatL}{\bimL}
\safemath{\rmatM}{\bimM}
\safemath{\rmatN}{\bimN}
\safemath{\rmatO}{\bimO}
\safemath{\rmatP}{\bimP}
\safemath{\rmatQ}{\bimQ}
\safemath{\rmatR}{\bimR}
\safemath{\rmatS}{\bimS}
\safemath{\rmatT}{\bimT}
\safemath{\rmatU}{\bimU}
\safemath{\rmatV}{\bimV}
\safemath{\rmatW}{\bimW}
\safemath{\rmatX}{\bimX}
\safemath{\rmatY}{\bimY}
\safemath{\rmatZ}{\bimZ}

\safemath{\rmatDelta}{\bimDelta}
\safemath{\rmatLambda}{\bimLambda}
\safemath{\rmatPhi}{\bimPhi}
\safemath{\rmatSigma}{\bimSigma}
\safemath{\rmatOmega}{\bimOmega}
\safemath{\rmatTheta}{\bimTheta}

\usepackage{amssymb}
\usepackage{amsfonts}
\usepackage{mathrsfs}
\usepackage{xspace}
\usepackage{bm}
\usepackage{fancyref}
\usepackage{textcomp}

\usepackage{multirow}
\usepackage{stmaryrd}

\newenvironment{textbmatrix}{	\setlength{\arraycolsep}{2.5pt}%
								\big[\begin{matrix}}{\end{matrix}\big]%
								\raisebox{0.08ex}{\vphantom{M}}}

\def\be{\begin{equation}}
\def\ee{\end{equation}}
\def\een{\nonumber \end{equation}}
\def\mat{\begin{bmatrix}}
\def\emat{\end{bmatrix}}
\def\btm{\begin{textbmatrix}}
\def\etm{\end{textbmatrix}}

\def\ba#1\ea{\begin{align}#1\end{align}}
\def\bas#1\eas{\begin{align*}#1\end{align*}}
\def\bs#1\es{\begin{split}#1\end{split}} 
\def\bg#1\eg{\begin{gather}#1\end{gather}}
\def\bml#1\eml{\begin{multline}#1\end{multline}}
\def\bi#1\ei{\begin{itemize}#1\end{itemize}}

\newcommand{\lefto}{\mathopen{}\left}

\DeclareMathOperator*{\argmin}{arg\;min}		% arg min
\DeclareMathOperator{\Exop}{\opE}			% expectation operator
\newcommand{\Ex}[2]{\ensuremath{\Exop_{#1}\lefto[#2\right]}} 	% expectation
\safemath{\dirac}{\delta}					% Dirac delta
\safemath{\krond}{\dirac}					% Kronecker delta
\safemath{\upto}{\uparrow}
\safemath{\downto}{\downarrow}
\safemath{\iu}{j}							% imaginary unit
\safemath{\ev}{\lambda}						% eigenvalue
\safemath{\hilseqspace}{l^{2}}				% Hilbert sequence space
\newcommand{\banachfunspace}[1]{\setL^{#1}}	% Banach function space
\safemath{\hilfunspace}{\banachfunspace{2}}	% Hilbert function space
\newcommand{\floor}[1]{\lfloor #1 \rfloor}

\safemath{\SNR}{\textsf{SNR}} 				% signal to noise ratio
\safemath{\PAR}{\textsf{PAR}} 				% signal to noise ratio
\safemath{\No}{N_0}							% noise spectral density
\safemath{\Es}{E_s}							% energy per symbol
\safemath{\Eb}{E_b}							% energy per bit
\safemath{\EbNo}{\frac{\Eb}{\No}}
\safemath{\EsNo}{\frac{\Es}{\No}}

\DeclareMathOperator{\CHop}{\ensuremath{\opH}} % channel operator
\safemath{\tvir}{\rndh_{\CHop}}				% time-varying impulse response
\safemath{\tvtf}{\rndl_{\CHop}}				% 	-''- transfer function
\safemath{\spf}{\rnds_{\CHop}}				% spreading function
\safemath{\bff}{H_{\CHop}}					% bi-freuqency function

\safemath{\ircf}{r_{h}}						% impulse response correlation fn.
\safemath{\tftvcf}{r_{s}}					% scattering function
\safemath{\tfcf}{r_{l}}						% time-frequency correlation fn.
\safemath{\bfcf}{r_{H}}						% bi-frequency correlation fn.

\safemath{\tcorr}{c_h}						% time-correlation function
\safemath{\scf}{c_{s}}						% spreading function
\safemath{\tfcorr}{c_{l}}					% transfer-function correlation
\safemath{\fcorr}{c_{H}}						% frequency-correlation function

\safemath{\mi}{I}							% mutual information
\safemath{\capacity}{C}						% capacity

\safemath{\normal}{\mathcal{N}}			% normal distribution
\safemath{\jpg}{\mathcal{CN}}			% jointly proper Gaussian
\safemath{\mchain}{\leftrightarrow}		% Markov chain
\safemath{\dB}{\,\mathrm{dB}}
\safemath{\dBm}{\,\mathrm{dBm}}
\safemath{\Hz}{\,\mathrm{Hz}}
\safemath{\kHz}{\,\mathrm{kHz}}
\safemath{\MHz}{\,\mathrm{MHz}}
\safemath{\GHz}{\,\mathrm{GHz}}
\safemath{\s}{\,\mathrm{s}}
\safemath{\ms}{\,\mathrm{ms}}
\safemath{\mus}{\,\mathrm{\text{\textmu}s}}
\safemath{\ns}{\,\mathrm{ns}}
\safemath{\ps}{\,\mathrm{ps}}
\safemath{\meter}{\,\mathrm{m}}
\safemath{\mm}{\,\mathrm{mm}}
\safemath{\cm}{\,\mathrm{cm}}
\safemath{\m}{\,\mathrm{m}}
\safemath{\W}{\,\mathrm{W}}
\safemath{\mW}{\, \mathrm{mW}}
\safemath{\J}{\,\mathrm{J}}
\safemath{\K}{\,\mathrm{K}}
\safemath{\bit}{\,\mathrm{bit}}
\safemath{\nat}{\,\mathrm{nat}}

\safemath{\define}{\triangleq}			% definition

\safemath{\equivalent}{\sim}
\safemath{\distas}{\sim}					% distributed according to
\safemath{\sdiff}{\Delta}				% symmetric set difference

\safemath{\reals}{\mathbb{R}}
\safemath{\positivereals}{\reals_{+}}
\safemath{\integers}{\mathbb{Z}}
\safemath{\posint}{\integers_{+}}
\safemath{\naturals}{\mathbb{N}}
\safemath{\posnaturals}{\naturals_{+}}
\safemath{\complexset}{\mathbb{C}}
\safemath{\rationals}{\mathbb{Q}}

\newcommand*{\fancyrefapplabelprefix}{app}		% Appendix
\newcommand*{\fancyrefthmlabelprefix}{thm}		% Theorem
\newcommand*{\fancyreflemlabelprefix}{lem}		% Lemma
\newcommand*{\fancyrefcorlabelprefix}{cor}		% Corollary
\newcommand*{\fancyrefdeflabelprefix}{def}		% Definition
\newcommand*{\fancyrefproplabelprefix}{prop}	% Proposition
\newcommand*{\fancyrefobslabelprefix}{obs}		% Observation 
\newcommand*{\fancyrefalglabelprefix}{alg}		% Algorithm
\newcommand*{\fancyrefasmlabelprefix}{asm}	    % Assumption
\newcommand*{\fancyrefasmslabelprefix}{asms}	    % Assumptions
\newcommand*{\fancyreftbllabelprefix}{tbl}	    % Table
\newcommand*{\fancyrefestilabelprefix}{esti}	    % Table

\frefformat{vario}{\fancyrefseclabelprefix}{Section~#1}
\frefformat{vario}{\fancyrefthmlabelprefix}{Theorem~#1}
\frefformat{vario}{\fancyreflemlabelprefix}{Lemma~#1}
\frefformat{vario}{\fancyrefcorlabelprefix}{Corollary~#1}
\frefformat{vario}{\fancyrefdeflabelprefix}{Definition~#1}
\frefformat{vario}{\fancyrefobslabelprefix}{Observation~#1}
\frefformat{vario}{\fancyrefasmlabelprefix}{Assumption~#1}
\frefformat{vario}{\fancyrefasmslabelprefix}{Assumptions~#1}
\frefformat{vario}{\fancyreffiglabelprefix}{Figure~#1}
\frefformat{vario}{\fancyrefapplabelprefix}{Appendix~#1} 
\frefformat{vario}{\fancyrefproplabelprefix}{Proposition~#1}
\frefformat{vario}{\fancyrefalglabelprefix}{Algorithm~#1}
\frefformat{vario}{\fancyrefeqlabelprefix}{(#1)}
\frefformat{vario}{\fancyreftbllabelprefix}{Table~#1}
\frefformat{vario}{\fancyrefestilabelprefix}{Estimator~#1}
\newtheorem{thm}{Theorem}
\newtheorem{cor}{Corollary}   % Turned off theorem numbering

\newtheorem{defi}{Definition}
\safemath{\dictab}{[\,\dicta\,\,\dictb\,]}

\safemath{\ysig}{\bmy}
\safemath{\ysighat}{\hat{\ysig}}
\safemath{\ysigdim}{M}
\safemath{\xsig}{\bmx}
\safemath{\xsigdim}{N}
\safemath{\nx}{n_x}
\safemath{\zsig}{\bmz}
\safemath{\zsigdim}{\ysigdim}
\safemath{\rsig}{\bmr}
\safemath{\Adict}{\bA}
\safemath{\Adicttilde}{\widetilde{\Adict}}
\safemath{\Adictdim}{\outputdim\times\xsigdim}
\safemath{\avec}{\bma}
\safemath{\avectilde}{\tilde{\avec}}
\safemath{\Bdict}{\bB}
\safemath{\Bdicttilde}{\widetilde{\Bdict}}
\safemath{\Cdict}{\bC}
\safemath{\cvec}{\bmc}
\safemath{\Ddict}{\bD}
\safemath{\Ddictdim}{\ysigdim\times\xsigdim}
\safemath{\dvec}{\bmd}
\safemath{\Ddicttilde}{\widetilde{\bD}}
\safemath{\Bonb}{\bB}
\safemath{\bvec}{\bmb}
\safemath{\Bonbdim}{\ysigdim\times\ysigdim}
\safemath{\noise}{\bmn}
\safemath{\noisedim}{\ysigim}
\safemath{\err}{\bme}
\safemath{\errdim}{\ysigdim}
\safemath{\errset}{\setE}
\safemath{\nerr}{n_e}
\safemath{\delop}{\bP_\errset}
\safemath{\delopc}{\bP_{{\errset}^c}}

\safemath{\cplxi}{\imath}
\safemath{\cplxj}{\jmath}
\safemath{\dict}{\matD}
\safemath{\inputdim}{N}		% number of columns of dictionary D
\safemath{\outputdim}{M}		%number of rows of dictionary D
\safemath{\sparsity}{S}	%sparsity
\safemath{\inputdimA}{{N_a}}	%total number of elements in dictionary A
\safemath{\inputdimB}{{N_b}}	%total number of elements in dictionary B
\safemath{\elemA}{{n_a}}	%number of elements chosen from dictionary A
\safemath{\elemB}{{n_b}}	%number of elements chosen from dictionary B
\safemath{\resA}{\matR_a}	%restriction map to elements of dictionary A
\safemath{\resB}{\matR_b}	%restriction map to elements of dictionary B
\safemath{\subD}{\matS} %subdictionary
\safemath{\subA}{\matS_a} %subdictionary part of A
\safemath{\subB}{\matS_b} %subdictionary part of B
\safemath{\dicta}{\matA} 	% first subdictionary
\safemath{\dictb}{\matB} 	% second subdictionary
\safemath{\hollowS}{H}
\safemath{\hollowA}{H_a}
\safemath{\hollowB}{H_b}
\safemath{\cross}{Z}
\safemath{\coh}{\mu_d}			% coherence dictionary
\safemath{\coha}{\mu_a}			% coherence first subdictionary
\safemath{\cohb}{\mu_b}			% coherence second subdictionary
\safemath{\mubs}{\nu}	%block sub-coherence
\safemath{\cohm}{\mu_m} %mutual coherence
\safemath{\dictset}{\setD}	% set of dictionaries
\safemath{\dictsetp}{\dictset(\coh,\coha,\cohb)}	% set of dictionaries parametrized
\safemath{\dictsetgen}{\dictset_\text{gen}}
\safemath{\dictsetgenp}{\dictsetgen(\coh)}
\safemath{\dictsetonb}{\dictset_\text{onb}}
\safemath{\dictsetonbp}{\dictsetonb(\coh)}

\safemath{\leftside}{U}
\safemath{\rightsideA}{R_a}
\safemath{\rightsideB}{R_b}

\safemath{\indexS}{\setI_S} %set of indices participating in sub-dictionary S

\safemath{\na}{n_a}			% cardinality of set of linearly independent columns of first dictionary
\safemath{\nb}{n_b}			% cardinality of set of linearly independent columns of second dictionary
\safemath{\coeffa}{p_i}	%coefficients for columns of A
\safemath{\coeffb}{q_j}	%coefficients for columns of B
\safemath{\seta}{\setP}		% set of linearly independent columns of A
\safemath{\setb}{\setQ}     % set of linearly independent columns of B
\safemath{\setw}{\setW}	%set of n largest elements of w
\safemath{\setz}{\setZ}	%set of L-n largest elements of z
\safemath{\cola}{\veca}		% generic element of the dictionary A
\safemath{\colb}{\vecb}		% generic element of the dictionary B
\safemath{\cold}{\vecd}		% generic element of the dictionary D
\safemath{\inputvec}{\vecx} 	%coefficient vector (input)
\safemath{\error}{\vece}	%error vector
\safemath{\noiseout}{\vecz} 	%noisy output vector
\safemath{\inputvecel}{x}
\safemath{\inputveca}{\vecx_a}
\safemath{\inputvecb}{\vecx_b}
\safemath{\outputvec}{\vecy}	%output of Dictionary
\safemath{\lambdamin}{\lambda_{\mathrm{min}}}
\safemath{\elltwo}{\ell_2}
\safemath{\ellone}{\ell_1}
\safemath{\ellzero}{\ell_0}
\safemath{\ellinf}{\ell_\infty}
\safemath{\ellinftilde}{\ell_{\widetilde\infty}}
\safemath{\licard}{Z(\coh,\coha,\cohb)}
\safemath{\xsol}{\hat{x}}
\safemath{\xbord}{x_b}		%Solution at the border
\safemath{\xstat}{x_s}		%Solution stationary in l0 prob
\safemath{\xstatLone}{\tilde{x}_s}
\safemath{\order}{\mathcal{O}} %order notation (big O)
\safemath{\scales}{\Theta} %scales as
\safemath{\ones}{\mathbf{1}} %all ones matrix
\safemath{\zeroes}{\mathbf{0}} %all zeroes matrix
\safemath{\thlone}{\kappa(\coh,\cohb)} %treshold l1 problem
\safemath{\constoneA}{\delta} %constant in l1 theorem to save space
\safemath{\constoneB}{\epsilon} %constant in l1 theorem to save space
\safemath{\nlarge}{L}				   %num large elements
\safemath{\sumlarge}{S_\nlarge}
\safemath{\maxlarger}{P_\nlarge}	   % maximum in Gribonval and Nielsen
\safemath{\Pzero}{\textrm{P0}}	
\safemath{\Pone}{\textrm{P1}}
\safemath{\vecfir}{\vecw}			 % \vecv element of the kernel of the dictionary, \vecv=[\vecfir \vecsec]
\safemath{\vecsec}{\vecz}
\safemath{\elvecfir}{w}              % element of vecfir
\safemath{\elvecsec}{z}				 % element of vecsec
\safemath{\nlargefir}{n}
\safemath{\normout}{\gamma}
\safemath{\auxfun}{h}
\safemath{\supp}{\textrm{supp}}%support

\safemath{\indexa}{\ell}
\safemath{\indexb}{r}
\safemath{\indexc}{i}
\safemath{\indexd}{j}

\safemath{\project}{P}%projector

\allowdisplaybreaks

\allowdisplaybreaks % THIS ALLOWS EQUATIONS TO BREAK THE PAGE
\IEEEoverridecommandlockouts

\usepackage{color}
\safemath{\Herm}{\textnormal{H}}
\safemath{\Tran}{\textnormal{T}}

\newcolumntype{P}[1]{>{\centering\arraybackslash}p{#1}}

\begin{document}

%\title{Initializing Neural Networks by Layer Fusion}
\title{MSE-Optimal Neural Network Initialization \\ via Layer Fusion}

\author{
\IEEEauthorblockN{Ramina Ghods$^{1}$, Andrew S. Lan$^{2}$, Tom Goldstein$^{3}$, and Christoph Studer$^{4}$}\\ \vspace{-0.1cm}
\IEEEauthorblockA{$^\text{1}$Carnegie Mellon University, Pittsburgh, PA; {rghods@cs.cmu.edu}} 
\IEEEauthorblockA{$^\text{2}$University of Massachusetts Amherst, Amherst, MA;  {andrewlan@cs.umass.edu}} 
\IEEEauthorblockA{$^\text{3}$University of Maryland, College Park, MD; {tomg@cs.umd.edu}} 
\IEEEauthorblockA{$^\text{4}$Cornell Tech, New York, NY; {studer@cornell.edu}} 
\thanks{The work of RG and CS was supported in part by Xilinx Inc.\ and by the US National Science Foundation under grants ECCS-1408006, CCF-1535897,  CCF-1652065, CNS-1717559, and ECCS-1824379.}
}

\maketitle

% ================================================================================
% ================================================================================
% ================================================================================
%%\vspace{-1.1cm}
\begin{abstract}
Deep neural networks achieve state-of-the-art performance for a range of classification and inference  tasks. However, the use of stochastic gradient descent combined with the nonconvexity of the underlying optimization problems renders parameter learning susceptible to initialization. To address this issue, a variety of  methods that rely on random parameter initialization or knowledge distillation have been proposed in the past. In this paper, we propose FuseInit, a novel method to initialize shallower networks by fusing neighboring layers of deeper networks that are trained with random initialization. We develop theoretical results and efficient algorithms for mean-square error (MSE)-optimal fusion of neighboring dense-dense, convolutional-dense, and convolutional-convolutional layers. We show experiments for a range of classification and regression datasets, which suggest that deeper neural networks are less sensitive to initialization and shallower networks can perform better (sometimes as well as their deeper counterparts) if initialized with FuseInit.%
\end{abstract}

% ================================================================================
% ================================================================================
% ================================================================================

%%%%%%%%%%%%%%%%%%%%%%%%%%%%%%%%%%%%%%%%%%%%%%%%%%%%%%%%%%

% !TEX root = 20CISS_FuseInit.tex
%%%%%%%%%%%%%%%%%%%%%%%%%%%%%%

\section{Introduction}
\label{sec:introduction}
 A prominent approach to improving the performance of artificial neural networks is to increase the number of network parameters~\cite{simonyan2014very,szegedy2015going}.
Theoretical and empirical evidence in \cite{allen2018convergence,zhang2016understanding,arora2018optimization} suggest that over-parametrization (more parameters in the network than in the training data) enables one to find better minimizers (and often faster) and reduce the generalization error. Furthermore, reference \cite{livni2014computational} has shown that finding global minimizers can be easier for sufficiently large networks. 

Unfortunately, the deployment of deep neural nets with a large number of parameters in resource-constrained systems, such as mobile devices, unmanned aerial vehicles, autonomous cars is extremely challenging in terms of both storage and computation \cite{howard2017mobilenets,sandler2018mobilenetv2}. 
Fortunately, the parameters of deep networks often exhibit high redundancy and, with appropriate initialization schemes, shallower networks can in many situations be trained to perform as well as their deeper counterparts~\cite{denil2013predicting,ba2014deep}. 
For example, reference~\cite{arora2018stronger} has demonstrated that one can substantially compress the number of parameters in deep networks, but training of such shallower networks directly, without using a deeper network, is a notoriously difficult task.
In many situations, the success or failure of training shallower networks depends on the initialization method---the design of powerful initialization strategies, however, remains an active research area.

 \captionsetup[figure]{textfont=it,font=normalsize}
\begin{figure*}[tp]
\centering
\captionsetup[subfigure]{textfont=it,font=small}
\begin{subfigure}[b]{0.5\columnwidth}
\centering 
\includegraphics[width=1\columnwidth]{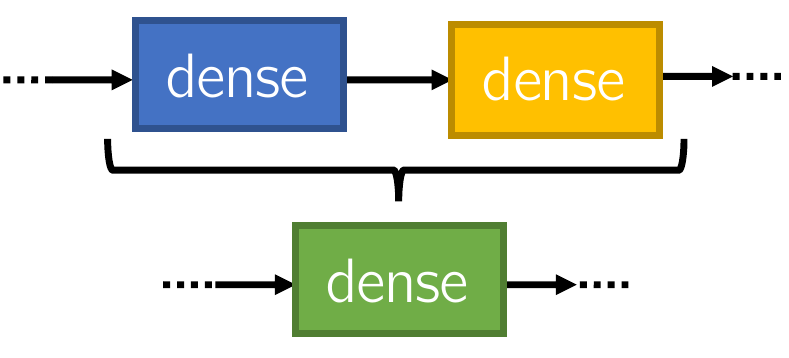}
\caption*{(a) dense-dense $\to$ dense}
\label{fig:cifar_he}
\end{subfigure} %\\[0.65cm]
\hfill
\begin{subfigure}[b]{0.5\columnwidth}
\centering 
\includegraphics[width=1\columnwidth]{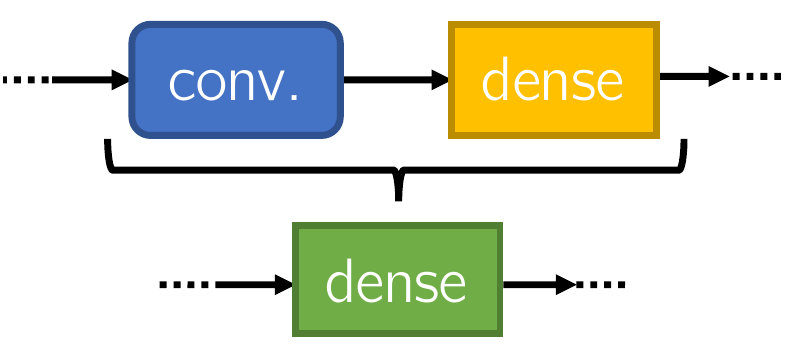}
\caption*{(b) conv.-dense $\to$ dense}
\label{fig:cifar_rnd}
\end{subfigure} %\\[0.65cm] 
\hfill
\begin{subfigure}[b]{0.5\columnwidth}
\centering 
\includegraphics[width=1\columnwidth]{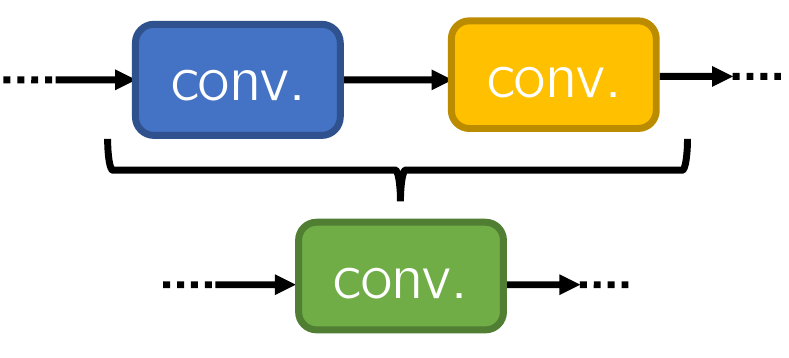}
\caption*{(c) conv.-conv. $\to$ conv.}
\label{fig:cifar_rnd}
\end{subfigure} %\\[0.65cm] 
\caption{The three considered scenarios of fusing neighboring dense and/or convolutional layers.}
\label{fig:fuse}
\vspace{-0.2cm}
\end{figure*}

%%%
\subsection{Contributions}
We propose FuseInit, a principled network initialization method.
The key idea of FuseInit is to first train a deeper neural network with  initialization methods that rely on random weights---the deeper network is then used to  initialize a shallower network by fusing neighboring layers.
Using a classical result by Bussgang~\cite{bussgang52a}, we develop new theory for mean-square error (MSE)-optimal fusion of neighboring dense-dense, convolutional-dense, convolutional-convolutional layers with arbitrary activation functions.
We propose efficient algorithms for FuseInit that scale favorably to deeper neural networks and large datasets.
To demonstrate the efficacy of our approach, we show experimental results for a range of classification and regression datasets. Our results suggest that deeper networks are less sensitive to initialization and shallower networks can perform better (sometimes as well as their deeper counterparts) if initialized with FuseInit.

%%%
\subsection{Relevant Prior Art}

The majority of parameter initialization schemes for neural nets deployed in practice rely on randomly initialized network parameters. 
A widespread approach to random initialization is the use of zero-mean Gaussian random variables with small variance (e.g., $0.01$)~\cite{krizhevsky2012imagenet}. 
Reference~\cite{glorot2010understanding} proposed random initialization with a variance that depends on the number of inputs and outputs of the layer to be initialized.
Reference~\cite{he2015delving} improved upon this approach for networks with ReLU activations by using random variables with variance~$2/N$, where $N$ stands for the number of inputs to the target layer.
Other methods that focus particularly on deep network initialization with random parameters have been proposed in, e.g.,  \cite{mishkin2015all,saxe2013exact}. Our focus is on initializing shallow networks.
FuseInit combines random initialization with an expansion-and-fusion strategy: To initialize a target network, first add one (or multiple) layers to the network, initialize the deeper network with random parameters, train it, and finally fuse it to the smaller target architecture.

A prominent approach to train shallow neural networks from deep networks is knowledge distillation~\cite{hinton2015distilling}. 
This approach builds upon the idea of imposing the outputs of a deeper teacher network to the outputs of the shallower student network.
FuseInit differs from such methods as it starts directly from a deeper network and successively fuses neighboring layers to initialize the parameters of the shallower network instead of training the shallower (student) network with the outputs of the deeper (teacher) network from scratch. 
FuseInit can be combined with such methods by initializing the student network, which can then be trained via knowledge distillation.

ExpandNet is a recent initialization method for shallow networks \cite{guo2019expandnets}.
The idea is to learn shallow nets by expanding each layer into multiple linear layers and training the expanded network. FuseInit differs from this approach in the following ways.
While ExpandNet is using linear layers, FuseInit is able to \emph{optimally} fuse \emph{nonlinear} layers. FuseInit also uses the MSE-optimal fusion weights as a starting point to retrain the shallower network. Our experiments indicate that this re-training step significantly improves the performance of the shallower network. 
While ExpandNet only relies on experiments, we provide theory for MSE-optimal fusion of neighboring layers and use experiments to demonstrate the effectiveness of FuseInit. We furthermore provide an MSE analysis for the fused layers, which provides a metric that can be used to determine which layers to fuse.

Slightly less related to FuseInit is the plethora of network simplification methods that aim at reducing the number of parameters of deep neural nets; see, e.g., \cite{han2015deep,guo2016dynamic} and the references therein. Pruning methods are among the most prominent ones and remove network parameters based on their magnitude \cite{han2015learning} or the cost function \cite{hassibi1993second,hanson1989comparing,lecun1990optimal}. 
Other network simplification methods include quantization \cite{gong2014compressing,courbariaux2015binaryconnect,gupta2015deep}, sparsity \cite{liu2015sparse}, and low-rank structure \cite{denton2014exploiting}. 
The concept of FuseInit can be generalized for a range of network architectures, including networks with sparse and low-rank structure.

% !TEX root = 20CISS_FuseInit.tex
%%%%%%%%%%%%%%%%%%%%%%%%%%%%%%

\section{FuseInit: MSE-Optimal Neural Network Initialization via Layer Fusion}
\label{sec:FuseInit}
We now detail FuseInit for the three cases illustrated in \fref{fig:fuse}: (a) Two dense layers are fused into one dense layer, (b) one convolutional layer and one dense layer are fused into one dense layer, and (c) two convolutional layers are fused into a convolutional layer.
We first summarize the notation and then present theoretical results for MSE-optimal fusion of neighboring layers.
Finally, we show an efficient FuseInit algorithm that scales to deep neural networks and large datasets.

\subsection{Notation}
Lowercase and uppercase boldface letters represent column vectors and matrices, respectively. For a matrix $\bA$, the transpose is  $\bA^T$, and the $i$th row and $j$th column entry is~$\bA[i,j]$. For a vector~$\bma$, the $i$th entry is~$\bma[i]$, and the sub-vector containing the $i$th to $j$th entries is $\bma[i:j]=\bma_{i:j}$; furthermore, $\bma[i:j:k]=\bma_{i:j:k}$ stands for a vector consisting of one entry every other $k$ entries taken from the $i$th to the $j$th entries of vector $\bma$;
$\sum_{i=1,i+=s}^{L} \bma[i]$ denotes summation of $\bma[i]$ starting from index $1$ to~$L$ with strides of $s$.
The $\ell_2$-norm of~$\bma$ is~$\|\bma\|_2$; $\text{flip}(\bma)$ denotes a vector $\bma$ with its entries in reverse order.

\subsection{FuseInit for Dense-Dense and Convolutional-Dense Layers}
\label{sec:densedense}
Consider the following model for two consecutive layers of a neural network, with $\bma_0 \in \reals^{L_0}$ as the input to the first layer and $\bma_2 \in \reals^{L_2}$ as the output of the second layer. Note that these can be any two neighboring layers in a deep neural network, as long as the second layer is a dense, fully-connected layer. As it will be clear later, there are no restrictions on the first layer since we only need its empirical moments. The function $H_1(\cdot)$ fully characterizes the input-output relation of the first layer.
Let the second layer use activation function $f_2(\cdot)$, weight matrix $\bW_2 \in \reals^{L_2 \times L_1}$, and bias vector $\bmb_2 \in \reals^{L_2}$.
The following model describes the end-to-end input-output relation of the two neighboring layers:
\begin{align}\label{eq:2layer_FC_model}
\bma_2 =f_2(\bW_2 \bma_1+\bmb_2) \quad \text{and} \quad \bma_1 =H_1(\bma_0).
\end{align}
Note that the inputs to the first and second layers may not be vectors; in this case, we vectorize $\bma_0$ and $\bma_1$.
In order to fuse two neighboring layers into one, we use the following three-step procedure. In the first step, we train the parameters of the entire network by random initialization using a standard training method, e.g., stochastic gradient descent. In the second step, we use the trained parameters to fuse the first and second layer into a single dense layer with input-output relation
\begin{align}\label{eq:equivalent_dense}
\bma_2=f_2(\tilde{\bW} \bma_0+\tilde{\bmb}),
\end{align}
where $\tilde{\bW}\in\reals^{L_2\times L_0}$ is a new weight matrix and $\tilde{\bmb}\in\reals^{L_2}$ a new bias vector; we keep the activation function~$f_2(\cdot)$ of the second layer. 
We select the new weight matrix and bias vector to minimize the MSE between the output of the initial two layers \fref{eq:2layer_FC_model} and the output of the new fused dense layer \fref{eq:equivalent_dense}.
Mathematically, we solve the following optimization problem:  
\begin{align}\label{eq:MSE_FC_minimization}
\{\tilde{\bW}^\star,\tilde{\bmb}^\star\}=\argmin_{\tilde{\bW}\in\reals^{L_2\times L_0},\tilde{\bmb}\in\reals^{L_2}} \, \textit{MSE}.
\end{align}
Here, the MSE is defined as
\begin{align}\label{eq:MSE_FC_expression}
 \textit{MSE} = \Ex{}{\big\|\big(\tilde{\bW} \bma_0+\tilde{\bmb}\big) - \left( \bW_2 H_1(\bma_0)+\bmb_2 \right)\big\|_2^2},
\end{align}
where the expectation $ \Ex{}{\cdot}$ is over the distribution of the input vector $\bma_0$.
In the third step, we retrain the entire fused neural network (including other layers) by initializing the fused layer with the new weight matrix $\tilde{\bW}^\star$ and new bias vector $\tilde{\bmb}^\star$ obtained from solving \fref{eq:MSE_FC_minimization}.
We note that while minimizing the MSE is not necessarily optimal in terms of classification or regression performance, it yields analytical expressions and efficient algorithms (see \fref{sec:practical}).

The following result for MSE-optimal weights and biases builds upon the nonlinear signal decomposition by J.~J.~Bussgang in~\cite{bussgang52a}. See \fref{app:linearize_FC} for the proof. 
\begin{thm}\label{thm:linearize_FC}
Let \fref{eq:2layer_FC_model} be the input-output relation of two neighboring layers of a trained neural net. Define the vectors $\overline{\bma}_0=\Ex{}{\bma_0} $ and $\overline{\bma}_1=\Ex{}{\bma_1}  = \Ex{}{H_1(\bma_0)}$, where expectation is over the distribution of~$\bma_0$. 
Define the covariance matrix 
\begin{align}
\bC_{\bma_0} = \Ex{}{(\bma_0-\overline\bma_0)(\bma_0-\overline\bma_0)^T},
\end{align}
and the cross-covariance matrix
\begin{align}
{\bC}_{\bma_1 \bma_0} = \Ex{\bma_0}{(\bma_1-\overline \bma_1)(\bma_0-\overline\bma_0)^T}\!.
\end{align}
By assuming that the covariance matrix $\bC_{\bma_0}$  is full rank, the new weight matrix $\tilde{\bW}^\star$ and bias vector~$\tilde{\bmb}^\star$ of the equivalent layer \fref{eq:equivalent_dense} that minimizes MSE in \fref{eq:MSE_FC_expression} are given by 
\begin{align}\label{eq:Wopt_FC}
\tilde{\bW}^\star \!= \bW_2 {\bC}_{\bma_1 \bma_0} \bC_{\bma_0}^{-1}  \,\text{ and } \,
\tilde{\bmb}^\star = \bW_2 \overline{\bma}_1 + \bmb_2 - \tilde{\bW}^\star\overline{\bma}_0.
\end{align}
\end{thm}

The only assumption required in \fref{thm:linearize_FC} is that the matrix~${\bC}_{\bma_0} $ has full rank; a more general condition is to use any new weight matrix $\tilde{\bW}^\star$ for which $\tilde{\bW}^\star \bC_{\bma_0} = \bW_2 {\bC}_{\bma_1 \bma_0}$. 
In our experiments with the algorithm detailed in \fref{sec:practical}, we have not observed this matrix to be rank deficient. 
Furthermore, we emphasize that the method in \fref{thm:linearize_FC} can also be used to fuse more than two consecutive layers and more general network structures---in this case, the function $H_1(\cdot)$ simply represents the effect of multiple layers. 

From \fref{thm:linearize_FC}, we can obtain the following compact expression for the MSE incurred by layer fusion; a short derivation is given in \fref{app:MSEexpression}.
\begin{cor} \label{cor:MSEexpression}
The MSE of the fused layer in \fref{eq:MSE_FC_expression} obtained by \fref{thm:linearize_FC} is given by
\begin{align} \label{eq:MSE}
\textit{MSE} =\mathrm{trace}\!\left( \bW_2\! \left(\bC_{\bma_1}-\bC_{\bma_1 \bma_0}\bC_{\bma_0}^{-1}\bC_{\bma_0 \bma_1}\right)\! \bW_2^T\right)\!.
\end{align}
\end{cor}
We note that this result can be used to determine which layers in a network to fuse. A detailed study on methods that select the best layers to fuse is left for future work.

\subsection{FuseInit of Convolutional-Convolutional Layers}
Consider the following model for two consecutive convolutional layers of a neural network. For the sake of simplicity, we detail the 1-dimensional case. 
The first layer has $M$ input channels, each of length~$L_0$, i.e., $\{\bma_0^1,\ldots,,\bma_0^{M}\}$, and $N$ output channels, each of length $L_1$, i.e., $\{\bma_1^1,\ldots,\bma_1^{N} \}$. 
The second layer has $P$ output channels, each of length~$L_2$, i.e., $\{\bma_2^1,\ldots,\bma_2^{P} \}$.
In what follows, we assume that the the following zero-padding strategy is implemented.
\begin{defi}
If the vector $\bmx$ is convolved with a filter of length $k$, then we pad the first and last entries of $\bmx$ with $\floor{\frac{k}{2}}$ and  $\floor{\frac{k-1}{2}}$ zeros, respectively. This zero-padding operation is denoted by~$\setZ^S(\bmx)$.
\end{defi}
The following model describes the input-output relation of the two neighboring convolutional layers:
\begin{align}\label{eq:2layer_generalconv_model}
\bma_1^{n}&= \textstyle f_1\big(\sum_{m=1}^{M} \bmh_1^{m,n} * \bma_0^{m} +\bmb_1^{n} \big), & n = 1,\ldots,N\\
\bma_2^{p}&=\textstyle f_2\big(\sum_{n=1}^{N} \bmh_2^{n,p} * \bma_1^{n} +\bmb_2^{p} \big), & p = 1,\ldots,P. \label{eq:2layer_generalconv_model1}
\end{align}
Here, the superscripts for the filters $\bmh_1^{m,n}$ and $\bmh_2^{n,p}$ denote the input and output channel index, respectively. 
We assume that the convolutions performed with the filters $\bmh_1^{m,n}$ and $\bmh_2^{n,p}$ have stride~$s_1$ and~$s_2$, respectively. The functions $f_1(\cdot)$ and $f_2(\cdot)$ describe each layer's activation function and a max-pool of stride $r_1$ and $r_2$; these functions can also represent batch normalization or dropout.

To fuse two neighboring convolutional layers into one, we use a three-step procedure similar to that in \fref{sec:densedense}.
In the first step, we train the parameters of the entire network using random initialization. 
In the second step, we use the trained parameters to fuse the two layers in~\fref{eq:2layer_generalconv_model} and \fref{eq:2layer_generalconv_model1} into a single convolutional layer with input-output relation:
\begin{align}\label{eq:equivalent_generalconv} 
\bma_2^{p}= \textstyle f_2\big(\sum_{m=1}^{M} \tilde{\bmh}^{m,p} * \bma_0^{m} +\tilde{\bmb}^{p} \big), \quad p = 1,\ldots,P.
\end{align}
Here, $\tilde{\bmh}^{m,p}$ are new filter coefficients and $\tilde{\bmb}^{p}$ new bias vectors; we keep the activation function $f_2(\cdot)$ of the second layer. 
Note that the convolution has stride $\tilde{s}$ and uses the same zero-padding strategy as defined above. 
As in \fref{sec:densedense}, we propose to select the new filter coefficients and bias vectors to minimize the MSE per output channel $p$ between the output of the initial two layers, denoted by~$\textit{C-MSE}^p$.
Put simply, we seek the quantities $\tilde{\bmh}^{m,p}$, $m=1,\ldots,M$, and $\tilde{\bmb}^{p}$ that minimize 
\begin{align}
&\textit{C-MSE}^p= \label{eq:MSE_generalconv} \\
&\textstyle \Ex{}{\!\left\|\!\left(\sum_{n=1}^{N} \bmh_2^{n,p} * \bma_1^{n} \!+\!\bmb_2^{p}\right) \!-\! \left(\sum_{m=1}^{M} \tilde{\bmh}^{m,p} * \bma_0^{m} \!+\!\tilde{\bmb}^{p}\right)\!\right\|_2^2}\!, \nonumber
\end{align}
for $p=1,\ldots,P,$ where expectation is over the distribution of the input vectors $\bma^m_0$, $m=1,\ldots,M$. 
In the third step, we retrain the entire fused neural net (including the other layers) by initializing the filters of the fused layer with the new filter coefficients and bias vectors obtained by minimizing \fref{eq:MSE_generalconv}.

We obtain the following result for MSE-optimal filters and bias vectors. The proof of the following theorem is provided in \fref{app:equivalent_generalconv}. In contrast to the proof in \fref{app:linearize_FC} for dense layers, the proof for convolutional layers is more involved considering that convolutional networks include input, output channels, and zero-padding. 
\begin{thm}\label{thm:equivalent_generalconv}
Let \fref{eq:2layer_generalconv_model} and \fref{eq:2layer_generalconv_model1} describe the input-output relation of two consecutive 1-dimensional convolutional layers of a trained deep neural network.  
Define $\overline{\bma}^m_0=\Ex{}{\bma^m_0}$, $m=1,\ldots,M$, and $\overline{\bma}^n_1=\Ex{}{\bma^n_1}$, $n=1,\ldots,N$. Furthermore, define the auxiliary quantities
\begin{align}
\bmv^p  &= \textstyle \sum_{n=1}^{N} \bmh_2^{n,p} * (\bma_1^n-\overline{\bma}_1^n), \\
\bmu^m  &=\mathrm{flip}[\setZ^s(\bma_0^m-\overline\bma_0^m)],
\end{align} 
and assume that input vectors $\bma_0^{m} $ from different channels $m$ are uncorrelated, i.e. 
\begin{align} \label{eq:uncorrelated}
\Ex{}{\big(\bma_0^{m}-\overline{\bma}_0^m\big)\big(\bma_0^{m'}-\overline{\bma}_0^{m'}\big)}=0 \quad \text{ for } \quad m \neq m'.
\end{align}
Select a new filter length $\tilde{k}$. Then, the filter and bias vectors that minimize \fref{eq:MSE_generalconv} of the convolutional layer in \fref{eq:equivalent_generalconv} for input and output channel indices $m'=1,\ldots,M$ and $p'=1,\ldots,P$ are given by 
	\begin{align}
	\tilde{\bmh}^{m',p'} &= \big({\bU^{m'}}\big)^{-1} \bmz^{m',p'}, \\\label{eq:bias_generalconv}
	\tilde{\bmb}^{p'} &= \textstyle \sum_{n=1}^{N} \bmh_2^{n,p'} * \overline\bma_1^{n} +\bmb_2^{p'} - \sum_{m=1}^{M} \tilde{\bmh}^{m,p'} * \overline\bma_0^{m}, 
	\end{align}
	with the two auxiliary quantities
	\begin{align}\label{eq:generalU}
	&\bU^{m'} \!\!= \! \textstyle \Ex{}{ \sum_{i=1,i+=\tilde s}^{L_0}  \! \left(\!\bmu^{m'}_{L_0-i+1:L_0-i+\tilde k}\!\right) \!\! \left(\!\bmu^{m'}_{L_0-i+1:L_0-i+\tilde k}\!\right)^T } \\\label{eq:generalV}
	&\bmz^{m',p'} =\textstyle  \Ex{}{  \sum_{i=1,i+=\tilde s}^{L_0}  \bmv^{p'}[\frac{i-1}{\tilde{s}}+1]  \,  \bmu^{m'}_{L_0-i+1:L_0-i+\tilde k}  }\!,
	\end{align}
	where the filter $\tilde{\bmh}$ has stride $\tilde{s} = s_1 r_1 s_2$. 
\end{thm}

Note that the above result requires the matrices ${\bU^{m'}}$ to be full rank; in all our experiments in \fref{sec:results}, we have not observed this matrix to be rank deficient.  
Furthermore, the assumption in~\fref{eq:uncorrelated} may not hold in practice, especially if the number of channels is large. In our experiments, however, different channels were  approximately uncorrelated. 
Similar to \fref{thm:linearize_FC}, the above result can be used to fuse multiple convolutional layers into one convolutional layer. In addition, a generalization to two or more dimensional convolutions follows analogously, but results in arduous expressions.
%

% !TEX root = 20CISS_FuseInit.tex
%%%%%%%%%%%%%%%%%%%%%%%%%%%%%%

\subsection{FuseInit in Practice}\label{sec:practical}
While the results in Theorems~\ref{thm:linearize_FC} and~\ref{thm:equivalent_generalconv} enable compact analytical expressions for MSE-optimal layer fusion, explicit results for the first and second moments are often unavailable. 
In fact, one would need to have knowledge of the data distribution. In addition, even if the distribution were known perfectly, analytically computing the first and second moment is often difficult, even for simple distributions. 
Since a vast amount of training data is available in most applications, we can replace the exact moments with empirical moments computed with training data. 
\fref{alg:fuse_FC} summarizes a practical approach to FuseInit for the case of fusing a neural net into a dense layer---the algorithm for fusing convolutional layers is analogous. 

\setlength{\textfloatsep}{2pt}% Remove \textfloatsep
\begin{algorithm}[ht]
	\caption{Practical FuseInit algorithm for fusing dense-dense and convolutional-dense layers}
	\label{alg:fuse_FC}
	Let the architecture in \fref{eq:2layer_FC_model} describe two consecutive fully-connected layers and let the assumptions in \fref{thm:linearize_FC} hold. 
	Then, FuseInit is given by the following 3-step process:
	\begin{enumerate}
		\item Train the original neural network using random initialization with $T$ training data samples.
%		 We name the trained parameters $\bW_1$,$\bmb_1^o$,$\bW_2^o$ and $\bmb_2^o$ for original
		\item Using the trained parameters, compute the fusion weight matrix $\tilde{\bW}^\star$ and bias vector $\tilde{\bmb}^\star$ in~\fref{eq:Wopt_FC}  by first and second empirical moments using the  $T$ training data samples.
		\item Replace the two fused layers in \fref{eq:2layer_FC_model} with the single dense layer $\bma_2=f_2(\tilde{\bW} \bma_0+\tilde{\bmb})$. Retrain the fused network by initializing the fused layer with $\tilde{\bW}^\star$ and $\tilde{\bmb}^\star$ and the remaining layers with the trained parameters from Step 1.

	\end{enumerate}
\end{algorithm}

In Step 1, one can use any of the existing random initialization methods. In our experiments, we will use zero-mean Gaussian random variables with variance $0.05$. Another widely used initializer is He-initializer in \cite{he2015delving}, where we sample from a truncated zero-mean Gaussian distribution with variance $2/N$ ($N$ is the number of inputs). We have excluded results for the He-initializer as they are indistinguishable to our current results.
In Step 2, we only need to sample $T$ vectors in the neural network that correspond to $T$ training samples to calculate the necessary empirical moments (which we all compute in parallel).  
As shown by \cite{vershynin2012close}, $T$ only needs to be slightly larger than the number of input dimensions of the layers ($L_0$ and $L_1$) for the empirical moments  to be accurate estimates of the true covariance matrices $\bC_{\bma_0}$ and $\bC_{\bma_1\bma_0}$. 
Hence, the computational complexity of FuseNet is dominated by neural network inference for the $T$ training samples and empirical computation of the two matrices $\bC_{\bma_0}$ and $\bC_{\bma_1\bma_0}$. 
In situations where the layers contain thousands of nodes, the inversion of~$\bC_{\bma_0}$ in \fref{eq:Wopt_FC} can be done implicitly using conjugate gradient methods. Furthermore, for such large networks, storage of~$\bC_{\bma_0}$ and $\bC_{\bma_1\bma_0}$ becomes the major bottleneck. 
In Step 3, the network is retrained using the same $T$ training samples. As we will show next, far fewer epochs are required to retrain the network to achieve good performance.

% !TEX root = 20CISS_FuseInit.tex
%%%%%%%%%%%%%%%%%%%%%%%%%%%%%%

%%%%%%%%%%%%%%%%%%%%%%%%%%%%%%%%%%%%%%%%%%%%%%%%%%%%	
%%%%%%%%%%%%%%%%%%%%%%%%%%%%%%%%%%%%%%%%%%%%%%%%%%%%	
%%%%%%%%%%%%%%%%%%%%%%%%%%%%%%%%%%%%%%%%%%%%%%%%%%%%	
% MASSIVE TABLE
%%%%%%%%%%%%%%%%%%%%%%%%%%%%%%%%%%%%%%%%%%%%%%%%%%%%	
%%%%%%%%%%%%%%%%%%%%%%%%%%%%%%%%%%%%%%%%%%%%%%%%%%%%	
%%%%%%%%%%%%%%%%%%%%%%%%%%%%%%%%%%%%%%%%%%%%%%%%%%%%	

\captionsetup[table]{textfont=it,font=normalsize}
\setlength{\textfloatsep}{10pt}% Remove \textfloatsep
\begin{table}[tp]
	\centering	
	%%%%%%%%%%%%%%%%%%%%%%%%%%%%%%%%%%%% 		
	\caption{Validation accuracy of convolutional-dense layers on CIFAR-10 dataset~\cite{krizhevsky2009learning}.}
	\vspace*{-0.1cm}
	\renewcommand{\arraystretch}{1.1}
	\scalebox{0.85}{ 
		\begin{tabular}{@{}p{4.14cm}P{2.42cm}P{2.42cm}P{2.42cm}P{2.42cm}@{}}
%			\hline
%			Initialization method & \multicolumn{2}{c|}{Random initializer} \\
			\toprule 
			\multirow{1}{*}{Algorithm} & \multicolumn{1}{c}{FuseInit} &  \multicolumn{1}{c}{Random}  \\
			%		 & Mean\,$\pm$\,SDV & Mean\,$\pm$\,SDV & Mean\,$\pm$\,SDV & Mean\,$\pm$\,SDV \\
			\midrule 
			6-layer: 32-32-64-64-128-128 &--&$0.8825\pm0.0040$\\
			%		\hline
			5-layer: 32-32-64-64-128 &$\textbf{0.8826}\pm0.0041$&$0.8691\pm0.0056$\\
			%		\hline
			4-layer: 32-32-64-64 &$\textbf{0.8535}\pm0.0046$&$0.8417\pm0.0060$\\ 
			\bottomrule
		\end{tabular}
	}
	\label{tbl:Conv2d_CIFAR10}
	
	%%%%%%%%%%%%%%%%%%%%%%%%%%%%%%%%%%%% 	
	\vspace*{0.5cm}
	\caption{Validation accuracy of convolutional-dense layers on Fashion-MNIST dataset~\cite{xiao2017fashion}.}
	\vspace*{-0.1cm}
	\renewcommand{\arraystretch}{1.1}
	\scalebox{0.85}{ 
		\begin{tabular}{@{}p{4.14cm}P{2.42cm}P{2.42cm}P{2.42cm}P{2.42cm}@{}}
%			\hline
%			Initialization method &\multicolumn{2}{c|}{Random initializer} \\
			\toprule 
			\multirow{1}{*}{Algorithm} & \multicolumn{1}{c}{FuseInit} &  \multicolumn{1}{c}{Random}  \\
			%		  & Mean\,$\pm$\,SDV & Mean\,$\pm$\,SDV & Mean\,$\pm$\,SDV & Mean\,$\pm$\,SDV \\
			\midrule 
			4-layer: 2-4-8-16 &--&$0.9107\pm 0.0024$\\
			%		\hline
			3-layer: 2-4-8 &$\textbf{0.9120}\pm 0.0025$&$0.9104\pm 0.0017$\\
			%		\hline
			2-layer: 2-4 &$\textbf{0.9010}\pm 0.0019$&$0.8971 \pm 0.0024$\\ 
			%		\hline
			1-layer: 2 &$\textbf{0.8803}\pm 0.0030$&$0.8756\pm 0.0043$\\ 
			\bottomrule
		\end{tabular}
	}
	\label{tbl:Conv2d_FashionMNIST}
	
	%%%%%%%%%%%%%%%%%%%%%%%%%%%%%%%%%%%% 	
		\vspace*{0.5cm}
	\caption{Validation accuracy of convolutional-convolutional layers on HAR dataset~\cite{anguita2013public}.}
	\vspace*{-0.1cm}
	\renewcommand{\arraystretch}{1.1}
	\scalebox{0.85}{ 		
		\begin{tabular}{@{}p{4.14cm}P{2.42cm}P{2.42cm}P{2.42cm}P{2.42cm}@{}}
%			\hline
%			Initialization method & \multicolumn{2}{c|}{Random initializer} \\
			\toprule
			\multirow{1}{*}{Algorithm}  & \multicolumn{1}{c}{FuseInit} &  \multicolumn{1}{c}{Random}  \\
			%		 & Mean\,$\pm$\,SDV & Mean\,$\pm$\,SDV & Mean\,$\pm$\,SDV & Mean\,$\pm$\,SDV \\
			\midrule
			2-layer: 18-36  & --&$0.962\pm0.002$\\
			%		\hline
			1-layer: 36    &$\textbf{0.958}\pm 0.005$&$\textbf{0.958} \pm 0.002$\\
			\bottomrule
		\end{tabular}	 
	}
	\label{tbl:Conv1d_HAR}
	
	%%%%%%%%%%%%%%%%%%%%%%%%%%%%%%%%%%%% 	
		\vspace*{0.5cm}
	\caption{Validation accuracy of convolutional-convolutional layers on speech commands dataset~\cite{warden2018speech}.}
	\vspace*{-0.1cm}
	\renewcommand{\arraystretch}{1.1}
	\scalebox{0.85}{ 		
		\begin{tabular}{@{}p{4.14cm}P{2.42cm}P{2.42cm}P{2.42cm}P{2.42cm}@{}}
%			\hline
%			Initialization method & \multicolumn{2}{c|}{Random initializer} \\
			\toprule 
			\multirow{1 }{*}{Algorithm}  & \multicolumn{1}{c}{FuseInit} &  \multicolumn{1}{c}{Random}  \\
			%			& Mean\,$\pm$\,SDV & Mean\,$\pm$\,SDV & Mean\,$\pm$\,SDV & Mean\,$\pm$\,SDV \\
			\midrule 
			4-layer: 32-32-64-64  & --& $0.887\pm 0.005$\\
			%		\hline
			3-layer: 32-64-64 &$\textbf{0.880}\pm 0.006$ &$0.868\pm 0.003$ \\
			\bottomrule
		\end{tabular}	 
	}
	\label{tbl:Conv1d_Speech}	
	
	%%%%%%%%%%%%%%%%%%%%%%%%%%%%%%%%%%%% 	
		
	\vspace*{0.5cm}
	\caption{Validation mean-absolute error (smaller is better) of dense-dense layers on wireless positioning dataset~\cite{studer2018channel}.}
	\vspace*{-0.1cm}
	\renewcommand{\arraystretch}{1.1}
	\scalebox{0.85}{ 	
		\begin{tabular}{@{}p{4.14cm}P{2.42cm}P{2.42cm}P{2.42cm}P{2.42cm}@{}}
			%			\hline
			%			Initialization method & \multicolumn{2}{c|}{Random initializer} \\
			\toprule 
			\multirow{1}{*}{Algorithm}  & \multicolumn{1}{c}{FuseInit} &  \multicolumn{1}{c}{Random}  \\
			%		 & Mean\,$\pm$\,SDV & Mean\,$\pm$\,SDV & Mean\,$\pm$\,SDV & Mean\,$\pm$\,SDV \\
			\midrule 
			3-layer: 16-128-2 & --&$7.426\pm0.112$\\
			%		\hline
			2-layer: 16-2 &$\textbf{7.221}\pm 0.336$&$7.277 \pm 0.472$\\
			%		\hline
			1-layer: 2 &$12.273 \pm 0.001$&$\textbf{12.262} \pm 0.0053$\\ 
			\bottomrule
		\end{tabular}
	}
	\label{tbl:Dense_Wireless}	
\end{table}

%\captionsetup[figure]{textfont=it,font=normalsize}
\begin{figure}[tp]
\centering
\captionsetup[subfigure]{textfont=it,font=small}
%
%
%\centering 
\includegraphics[width=0.850\columnwidth]{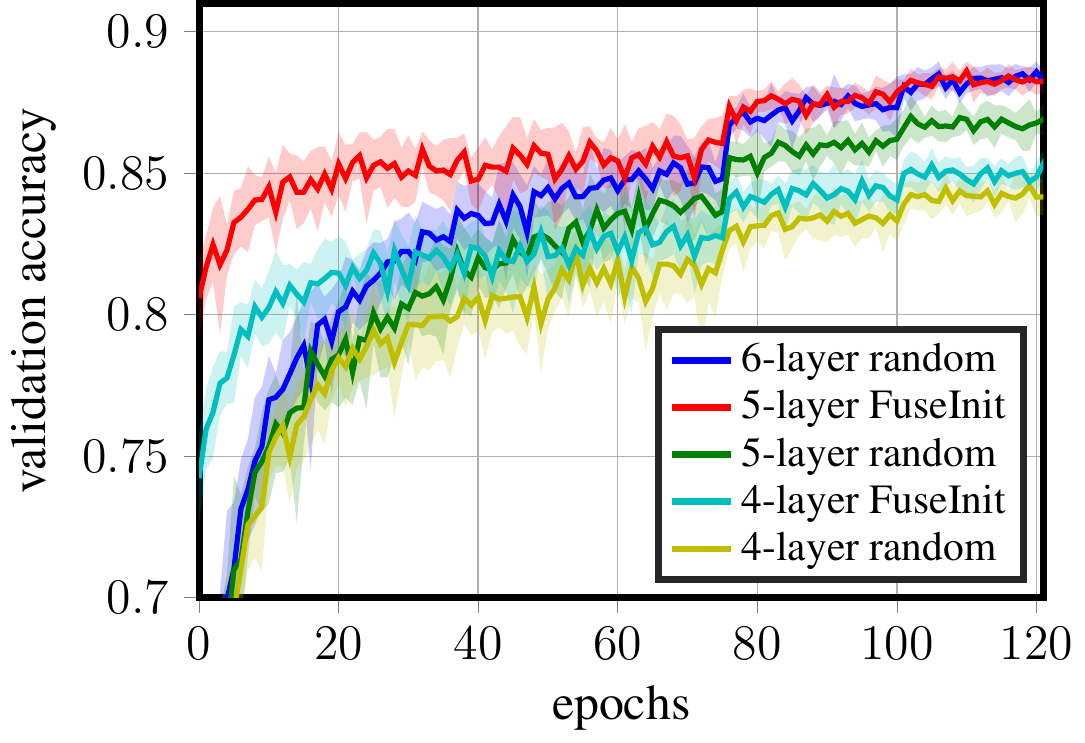}
%\caption{CIFAR-10 with random initializer}
%\label{fig:cifar_rnd}
%\end{subfigure} %\\[0.65cm] 
%
\caption{Comparison of validation accuracy for FuseInit and random initialization for different convolutional nets on CIFAR-10. FuseInit outperforms random initialization for the considered scenario; the 5-layer FuseInit net achieves the same accuracy as the randomly-initialized 6-layer net.}
\label{fig:cifar10}
\vspace{-0.1cm}
\end{figure}

\newcommand{\rom}[1]{\uppercase\expandafter{\romannumeral #1\relax}}

\section{Experimental Results}
\label{sec:results}

We now demonstrate the efficacy of FuseInit on five datasets. Tables \rom{1} to \rom{5} summarize the validation accuracy (or loss) of FuseInit on CIFAR-10 \cite{krizhevsky2009learning}, Fashion-MNIST~\cite{xiao2017fashion}, human activity recognition (HAR) \cite{anguita2013public}, speech commands \cite{warden2018speech} and wireless positioning \cite{studer2018channel} dataset. 
For each row of each table, we fuse one-by-one the layers of the network using FuseInit. We then report the mean and standard deviation of the achieved validation accuracy (or loss) over 10 trials in comparison to a randomly initialized network. The left column lists the number of nodes (channels) used per layer of the corresponding dense (convolutional) network.
Furthermore, we carry out a sufficiently large number of epochs for all experiments so that the validation accuracy (or loss) settles to a stable value.

To further illustrate the efficacy of FuseInit, we provide \fref{fig:cifar10}. This figure shows the mean and standard deviation of the validation accuracy over training epochs for CIFAR-10. Clearly, FuseInit provides a high-quality starting point for the network parameters,
which helps the network to converge to an accuracy that is superior to that of randomly-initialized networks with the same topology.  (The accuracy jump at epoch 75 is due to reduction of learning rate which is used to improve performance.)
Overall, our results indicate that neural networks that are initialized with FuseInit perform better that their randomly initialized counterparts.

% !TEX root = 20CISS_FuseInit.tex
%%%%%%%%%%%%%%%%%%%%%%%%%%%%%%%%%%%%%%%%%%%%%%%%%%%%%%%%%%%%
%
\section{Conclusions}
We have proposed FuseInit, a novel method to fuse neighboring layers in multi-layer neural networks.
FuseInit can be used to initialize shallower networks by first training deeper dense or convolutional networks with random weight initialization strategies, followed by layer fusion and retraining. 
For MSE-optimal layer fusion, we have developed analytical results and efficient algorithms.
Our experiments on five datasets have shown that FuseInit is able to consistently outperform random weight initialization methods. Furthermore, our results reveal that shallower networks can sometimes perform as well as their deeper counterparts if initialized with FuseInit.

There are many avenues for future work. FuseInit can be modified to train and initialize networks with special structure, such as residual or sparse networks---a corresponding study is part of ongoing work. 
The MSE expression in \fref{cor:MSEexpression} can potentially be used to identify the best layers that should be fused in deep network architectures. 
Furthermore, since FuseInit builds upon ideas from Bussgang's theorem, one could study lower-bounds on the information flow of neural networks.

\appendices

% !TEX root = 20CISS_FuseInit.tex
%%%%%%%%%%%%%%%%%%%%%%%%%%%%%%
%

\section{Proof of \fref{thm:linearize_FC}}
\label{app:linearize_FC}
%\begin{align*}
%\argmin_{\tilde{\bF_1},\tilde{\bmd_1}}\Ex{}{\|\bma_1-(\tilde\bF_1 \bma_0+\tilde\bmd_1)\|_2^2}
%\end{align*}
We wish to minimize post-fusion MSE in \fref{eq:MSE_FC_expression}. 
Our approach builds upon a generalization of the nonlinear, scalar signal decomposition by \cite{bussgang52a} to an affine vector  decomposition. Specifically, we first compute the new MSE-optimal bias vector~$\tilde{\bmb}^\star$. Since \fref{eq:MSE_FC_expression} is a quadratic form, we can take the derivative with respect to~$\tilde{\bmb}$ and setting it to zero, which yields  
\begin{align}
\frac{\partial}{\partial \tilde{\bmb}}\Ex{}{\big\|\big(\tilde{\bW} \bma_0+\tilde{\bmb}\big) - \left( \bW_2 \bma_1+\bmb_2 \right)\big\|_2^2}\! & = 0 \\
\frac{\partial}{\partial \tilde{\bmb}} \Ex{}{ \|\tilde{\bmb}\|^2_2+2\tilde{\bmb}^T\left(\tilde{\bW} \bma_0-(\bW_2 \bma_1+\bmb_2 ) \right) }\! & =0.
\end{align}
Here, expectation is over the distribution of the input data $\bma_0$. 
Basic matrix-vector calculus yields 
\begin{align}
\tilde{\bmb}^\star = \bW_2 \overline{\bma}_1+\bmb_2 - \tilde{\bW} \overline{\bma}_0,
\end{align}
where $\overline \bma_1=\Ex{}{\bma_1}  = \Ex{\bma_0}{H_1(\bma_0)}$ and $\overline \bma_0 = \Ex{}{\bma_0}$.
Next, we replace $\tilde{\bmb} $ in MSE expression and take the derivative with respect to the new weight matrix $\tilde{\bW}$ and set it to zero:
\begin{align}
\frac{\partial}{\partial \tilde{\bW}}&\Ex{}{\big\|\tilde{\bW} \left(\bma_0 - \overline{\bma}_0\right) - \bW_2 \left( \bma_1 - \overline{\bma}_1 \right)\big\|_2^2} \nonumber \\
&= \Exop{}\left[ \frac{\partial}{\partial \tilde{\bW}} \|\tilde{\bW} \left(\bma_0 - \overline{\bma}_0\right)\|_2^2   + \|  \bW_2 \left( \bma_1 - \overline{\bma}_1 \right) \|_2^2 \right. \nonumber\\
&\left. -  2  \left( \bma_1 - \overline{\bma}_1 \right) ^T \bW_2^T \tilde{\bW}  \left(\bma_0 - \overline{\bma}_0\right)  \right]\\
&= \tilde{\bW} \bC_{\bma_0} - \bW_2 \bC_{\bma_1 \bma_0}  =  0. \label{eq:importantcondition}
\end{align}
This expression results in the one provided in \fref{eq:Wopt_FC}. Note that even if  $\bC_{\bma_0} $ is not invertible, the result in~\fref{eq:importantcondition} can be used to find an MSE-optimal weight matrix by computing a matrix $\tilde{\bW} $ that satisfies the following condition:
\begin{align}
\tilde{\bW} \bC_{\bma_0} = \bW_2 \bC_{\bma_1 \bma_0}.
\end{align}

\section{Proof of \fref{cor:MSEexpression}}
\label{app:MSEexpression}
As an immediate consequence of Bussgang's decomposition in \cite{bussgang52a}, and with the optimal quantities $\tilde{\bW}^\star$ and $\tilde{\bmb}$ obtained above, the MSE in \fref{eq:MSE_FC_expression} can be expressed as follows:
\begin{align}
\textit{MSE} &= \Exop{} \left[ \|\tilde{\bW} \left(\bma_0 - \overline{\bma}_0\right)\|_2^2   + \|  \bW_2 \left( \bma_1 - \overline{\bma}_1 \right) \|_2^2  \right]
\nonumber\\
	&\,\,\,\,\,\,\,\left. -  2  \left( \bma_1 - \overline{\bma}_1 \right) ^T \bW_2^T \tilde{\bW}  \left(\bma_0 - \overline{\bma}_0\right) \right]\\
& =  \mathrm{trace} \!\left(  \tilde{\bW} \bC_{\bma_0} \tilde{\bW} ^T +   \bW_2  \bC_{\bma_1}  \bW_2^T  - 2  \tilde{\bW} \bC_{\bma_0 \bma_1} \bW_2^T \right)\\
%&= \mathrm{trace} \!\left( \bW_2 \bC_{\bma_1 \bma_0} \bC_{\bma_0}^{-1} \bC_{\bma_0} \bC_{\bma_0}^{-T} \bC_{\bma_0 \bma_1} \bW_2^T + \bW_2  \bC_{\bma_1}  \bW_2^T  -2  \bW_2 \bC_{\bma_1 \bma_0} \bC_{\bma_0}^{-1}  \bC_{\bma_0 \bma_1} \bW_2^T   \right)\\
&= \mathrm{trace} \!\left( \bW_2 \bC_{\bma_1 \bma_0}  \bC_{\bma_0}^{-T} \bC_{\bma_0 \bma_1} \bW_2^T + \bW_2  \bC_{\bma_1}  \bW_2^T  \right.\nonumber\\
&\,\,\,\,\,\,\, \left. -2  \bW_2 \bC_{\bma_1 \bma_0} \bC_{\bma_0}^{-1}  \bC_{\bma_0 \bma_1} \bW_2^T   \right)\\
&= \mathrm{trace} \!\left( \bW_2 \left( \bC_{\bma_1} - \bC_{\bma_1 \bma_0} \bC_{\bma_0}^{-1}  \bC_{\bma_0 \bma_1} \right) \bW_2^T \right)\!.
\end{align}
Note that this expression requires invertibility of $\bC_{\bma_1}$.

\section{Proof of \fref{thm:equivalent_generalconv}}
\label{app:equivalent_generalconv}

In contrast to the simple derivation in \fref{app:linearize_FC}, the proof for convolutional networks is more cumbersome. 
We first compute the MSE-optimal bias vector~$\tilde\bmb^{p'}$ by taking the derivative of the quantity $\textit{C-MSE}^{p'}$ in~$\tilde\bmb^{p'}$ and setting it to zero, i.e.,  
\begin{align}\nonumber
 \frac{\partial}{\partial\tilde\bmb^{p'}}  \Exop{} \Bigg[ \Bigg \| \textstyle & \left(\sum_{n=1}^{N} \bmh_2^{n,p'} * \bma_1^{n} +\bmb_2^{p'}\right) \\
 	&\left. -\left(\sum_{m=1}^{M} \tilde{\bmh}^{m,p'} * \bma_0^{m} +\tilde{\bmb}_2^{p'}\right)\right\|_2^2 \Bigg]\!= 0 ,
\end{align}
which, with basic matrix-vector calculus, yields the expression in \fref{eq:bias_generalconv}.
We next replace the new bias vector $\tilde\bmb^{p'}$ in $\textit{C-MSE}^{p'}$ to obtain
\begin{align}\nonumber
\textit{C-MSE}^{p'}= \textstyle 
\Exop{}\| \sum_{n=1}^{N} &\bmh_2^{n,p'} * \left(\bma_1^{n} -\overline{\bma_1^n} \right) \\
&- \sum_{m=1}^{M} \tilde{\bmh}^{m,p'} * \left( \bma_0^{m} - \overline{\bma_0^m}\right)\|_2^2\!
\end{align}
Our next objective is to solve for the new filter vectors $\tilde\bmh^{m',p'}$ for any given $m'$ and $p'$. 
To this end, we take the derivative of $\textit{C-MSE}^{p'}$ in $\tilde\bmh^{m',p'}$. By defining $\bmv^{p'} = \sum_{n=1}^{N} \bmh_2^{n,p'} * \left(\bma_1^{n}-\overline{\bma}_1^n \right)$, the derivative of $\textit{C-MSE}^{p'}$ in $\tilde\bmh^{m',p'}$ simplifies to the following expression:
\begin{align}\nonumber
&\frac{\partial}{\partial\tilde\bmh^{m',p'}} \textit{C-MSE}^{p'}  = \\
& \, 
 \frac{\partial}{\partial\tilde\bmh^{m',p'}}\textstyle  \Ex{}{ \| \sum_{m=1}^{M} \tilde{\bmh}^{m,p'} * \left( \bma_0^{m} - \overline{\bma}_0^m\right)\|_2^2 } \label{eq:term1_general} \\
&-2 \frac{\partial}{\partial\tilde\bmh^{m',p'}} \Ex{}{ {\bmv^{p'}}^T \textstyle \left( \sum_{m=1}^{M} \tilde{\bmh}^{m,p'} * \left( \bma_0^{m} - \overline{\bma}_0^m\right)\right)}\!. \label{eq:term2_general}
\end{align}
In order to compute the convolution $\tilde{\bmh}^{m,p'} * (\bma_0^m- \overline{\bma}_0^m)$ for $m=1,\ldots,M$, we need to zero-pad the vector $(\bma_0^m- \overline{\bma}_0^m)$. We call the zero-padded vector $\bmz_0^m=\setZ^s(\bma_0^m-\overline{\bma}_0^m)$. To compute the convolution, we slide the filter vector $\tilde{\bmh}^{m,p'}$ over $\bmz_0^m$ and compute the resulting inner products. Hence, if the convolution had a stride of 1, then the $i$th element of the convolution result $\tilde{\bmh}^{m,p'} * (\bma_0^m- \overline{\bma}_0^m)$ is given by
\begin{align}
\bmz_0[i:i+\tilde k-1]^T\text{flip}(\tilde{\bmh}),
\end{align}
or, equivalently,
\begin{align}\nonumber
&\left(\text{flip}(\bmz_0^m[i:i+\tilde k-1])\right)^T \tilde{\bmh}\\
&=\left(\bmu^m[L_0-i+1:L_0-i+\tilde k]\right)^T \tilde{\bmh}. \label{eq:tildegeneralconv_element}
\end{align}
Here, we define $\bmu^m=\text{flip}(\bmz_0^m)$.
With a stride of $\tilde{s}$, we only keep every $\tilde{s}$th element, i.e., we retain the indices $\ell \in [1:L_0:\tilde{s}]$.

We can now compute the derivative of $\textit{C-MSE}^{p'}$. We begin by the first term in  \fref{eq:term1_general}, which yields
\begin{align}
& \frac{\partial}{\partial\tilde\bmh^{m',p'}}  \textstyle \Ex{}{ \| \sum_{m=1}^{M} \tilde{\bmh}^{m,p'} * \left( \bma_0^{m} - \overline{\bma}_0^m\right)\|_2^2 }\\
&=\, \frac{\partial}{\partial\tilde\bmh^{m',p'}} \textstyle \Ex{}{ \| \tilde{\bmh}^{m',p'} * \left( \bma_0^{m'} - \overline{\bma}_0^{m'}\right)\|_2^2 }\\\label{eq:two}
&+\frac{\partial}{\partial\tilde\bmh^{m',p'}} \textstyle \Ex{}{ \| \sum_{m=1,m \neq m'}^{M} \tilde{\bmh}^{m,p'} * \left( \bma_0^{m} - \overline{\bma}_0^m\right)\|_2^2 }\\\nonumber
\label{eq:three}&+ \frac{\partial}{\partial\tilde\bmh^{m',p'}} \textstyle \Exop{} \Big[\sum_{m=1,m \neq m'}^{M} \left(\tilde{\bmh}^{m',p'} * ( \bma_0^{m'} - \overline{\bma}_0^{m'})\right)^T\\
&\hspace{4cm}\left(\tilde{\bmh}^{m,p'} * \left( \bma_0^{m} - \overline{\bma}_0^m\right)\right) \Big]\\\label{eq:simplified_term1}
&\stackrel{(a)}{=} \, \frac{\partial}{\partial\tilde\bmh^{m',p'}} \Ex{}{ \| \tilde{\bmh}^{m',p'} * \left( \bma_0^{m'} - \overline{\bma}_0^{m'}\right)\|_2^2 }.
\end{align}
Here, Step $(a)$ comes from the fact that both \fref{eq:two} and \fref{eq:three} are zero. Equation \fref{eq:two} is zero as it has no dependency on the derivative in $\tilde\bmh^{m',p'}$. To see why \fref{eq:three} is zero, we can expand this expression using the convolution result in \fref{eq:tildegeneralconv_element}:
\begin{align}
&\hspace{-0.3cm} \textstyle \Ex{}{ \sum_{m=1,m \neq m'}^{M} \left(\tilde{\bmh}^{m',p'} * ( \bma_0^{m'} - \overline{\bma}_0^{m'})\right)^T \tilde{\bmh}^{m,p'} * \left( \bma_0^{m} - \overline{\bma}_0^m\right) }\\\nonumber
& \textstyle =\Exop{} \Big[\sum_{m=1,m \neq m'}^{M} \sum_{i=1,i+=\tilde s}^{L_0} {\left(\bmu^{m'}_{L_0-i+1:L_0-i+\tilde k}\right)}^T \\ 
&\hspace*{2.5cm} \tilde{\bmh}^{m',p'}  {\left(\tilde{\bmh}^{m,p'}\right)}^T \!\!\!\left(\bmu^{m}_{L_0-i+1:L_0-i+\tilde k}\right)\Big]\\\nonumber
& \textstyle \stackrel{(b)}{=} \sum_{m=1,m \neq m'}^{M}  \sum_{i=1,i+=\tilde s}^{L_0}   \mathrm{trace}\Bigg(  \! \!  {\left(\tilde{\bmh}^{m,p'}\right)}^T \\
& \,\,\,\,  \Exop{} \! \Big[ \left(\bmu^{m}_{L_0-i+1:L_0-i+\tilde k}\right)  \!\!  {\left(\bmu^{m'}_{L_0-i+1:L_0-i+\tilde k}\right)}^T \Big]  \tilde{\bmh}^{m',p'}    \!\!  \Bigg), \label{eq:expressionthatiszero}
\end{align}
where Step $(b)$ follows from the fact that we can cyclically exchange terms under the trace operator. 
Using the assumption that the input vectors $\bma_0^{m'} $ and $\bma_0^{m} $ from different channels $m'\neq m$ are uncorrelated, i.e., 
\begin{align}
\Ex{}{   ( \bma_0^{m} - \overline{\bma}_0^{m})  ( \bma_0^{m'} - \overline{\bma}_0^{m'})^T }& =\bZero
\end{align}
it follows that 
\begin{align}
\Ex{}{ \left(\bmu^{m}_{L_0-i+1:L_0-i+\tilde k}\right)   {\left(\bmu^{m'}_{L_0-i+1:L_0-i+\tilde k}\right)}^T }& =\bZero,
\end{align}
which causes the expression in \fref{eq:expressionthatiszero} to be zero.
While this assumption may be violated in practice, it is reasonable to assume that carefully-designed filter outputs should be uncorrelated, because otherwise the filter channels would exhibit redundancy. 
Furthermore, by using this assumption within FuseInit, we implicitly learn new filter channels that promote this property.
We are now able to obtain a simple expression of the derivative in \fref{eq:simplified_term1} as follows:
\begin{align}
&\frac{\partial}{\partial\tilde\bmh^{m',p'}} \Ex{}{ \| \tilde{\bmh}^{m',p'} * \left( \bma_0^{m'} - \overline{\bma}_0^{m'}\right)\|_2^2 }\\\nonumber
&  \stackrel{(c)}{=} \Exop{}\Bigg[  \frac{\partial}{\partial\tilde\bmh^{m',p'}}  \textstyle  \sum_{i=1,i+=\tilde s}^{L_0}  \\ 
&\hspace*{1.2cm} \left| \left(\bmu^{m'}[L_0-i+1:L_0-i+\tilde k]\right)^T \tilde{\bmh}^{m',p'}    \right|_2^2   \Bigg]\\\nonumber
& \textstyle =  2 \Exop{}\Bigg[  \sum_{i=1,i+=\tilde s}^{L_0}   \left(\bmu^{m'}_{L_0-i+1:L_0-i+\tilde k}\right)\! \\
&\hspace*{3.5cm} \left(\bmu^{m'}_{L_0-i+1:L_0-i+\tilde k}\right)^T \Bigg]  \tilde{\bmh}^{m',p'}.
\end{align}
Here, Step $(c)$ follows from the convolution result in \fref{eq:tildegeneralconv_element}.

Next, we compute the second term in \fref{eq:term2_general}, which yields
\begin{align}
&-2 \frac{\partial}{\partial\tilde\bmh^{m',p'}} \textstyle \Ex{}{ {\bmv^{p'}}^T \left( \sum_{m=1}^{M} \tilde{\bmh}^{m,p'} * \left( \bma_0^{m} - \overline{\bma_0^m}\right)\right)}\\\nonumber
&  \stackrel{(d)}{=} -2 \Exop{}\big[\frac{\partial}{\partial\tilde\bmh^{m',p'}} \textstyle \sum_{m=1}^{M} \sum_{i=1,i+=\tilde s}^{L_0}   \bmv^{p'}[\frac{i-1}{\tilde{s}}+1]   \\  &\hspace*{3cm}\left(  \left(\bmu^m_{L_0-i+1:L_0-i+\tilde k}\right)^T \tilde{\bmh}^{m,p'}  \right) \big]\\
& \textstyle = -2 \Ex{}{  \sum_{i=1,i+=\tilde s}^{L_0}  \bmv^{p'}[\frac{i-1}{\tilde{s}}+1]  \,\,\,  \bmu^{m'}_{L_0-i+1:L_0-i+\tilde k}  }\!.
\end{align}
Here, Step $(d)$ follows from expanding the inner product between $\bmv^{p'}$ and $\sum_{m=1}^{M} \tilde{\bmh}^{m,p'} * \left( \bma_0^{m} - \overline{\bma_0^m}\right)$ and by using the convolution result in~\fref{eq:tildegeneralconv_element}.
Finally, by summing \fref{eq:term1_general} and \fref{eq:term2_general} and setting it to zero, we obtain
\begin{align}
& \textstyle 2 \Ex{}{  \sum_{i=1,i+=\tilde s}^{L_0}   \left(\bmu^{m'}_{L_0-i+1:L_0-i+\tilde k}\right) \!\!\! \left(\bmu^{m'}_{L_0-i+1:L_0-i+\tilde k}\right)\!^T } \!\! \tilde{\bmh}^{m',p'}  \nonumber \\
& \textstyle  \qquad -2 \Ex{}{  \sum_{i=1,i+=\tilde s}^{L_0}  \bmv^{p'}\left[\frac{i-1}{\tilde{s}}+1\right]  \bmu^{m'}_{L_0-i+1:L_0-i+\tilde k}  } = 0,
\end{align}
or equivalently
\begin{align}
\tilde{\bmh}^{m',p'} = \bU^{-1} \bV,
\end{align}
where $\bU$ and $\bV$ are defined in \fref{eq:generalU} and \fref{eq:generalV}.

%%%%%%%%%%%%%%%%%%%%%%%%%%%%%%%%%%%%%%%%%%%%%%%%%%%%%%%%%%

% ================================================================================
% ================================================================================
% ================================================================================

\section*{Acknowledgments}
The authors thank S. Jacobsson and G. Durisi for crucial discussions and insights on Bussgang's decomposition. 

% ================================================================================
% ================================================================================
% ================================================================================

\balance

%\vspace{0.2cm}
\bibliographystyle{IEEEtran} 
\bibliography{confs-jrnls,publishers,NNcompression}

%\balance

\end{document}